\documentclass[conference,a4paper]{IEEEtran}
\usepackage{cite}
\usepackage{amsmath,amssymb,amsfonts}
\usepackage{graphicx}
\usepackage{url}
\usepackage{subcaption}

\usepackage{textcomp}
\usepackage{xcolor}
\def\BibTeX{{\rm B\kern-.05em{\sc i\kern-.025em b}\kern-.08em
    T\kern-.1667em\lower.7ex\hbox{E}\kern-.125emX}}
\begin{document}

\title{Joint Architecture-Token-Bitwidth Multi-Axis Optimization of Vision Transformers for Semiconductor IC Packaging}


\author{
\IEEEauthorblockN{
Phat Nguyen\IEEEauthorrefmark{1},
Xue Geng\IEEEauthorrefmark{2},
Kaixin Xu\IEEEauthorrefmark{2},
Wang Zhe\IEEEauthorrefmark{2},
Xulei Yang\IEEEauthorrefmark{2},
Ngai-Man Cheung\IEEEauthorrefmark{1}
}

\IEEEauthorblockN{
\IEEEauthorrefmark{1}Temasek Laboratories, Singapore University of Technology and Design (SUTD) \\
E-mail: \{tienphat\_nguyen,ngaiman\_cheung\}@sutd.edu.sg
}
\IEEEauthorblockN{
\IEEEauthorrefmark{2}Agency for Science, Technology,
and Research (A*STAR) \\
E-mail: \{geng\_xue, xu\_kaixin, zhe\_wang, yang\_xulei\}@a-star.edu.sg}
}

\maketitle

\begin{abstract}
Vision Transformers (ViTs) have achieved strong performance in visual recognition, yet their deployment in resource-constrained industrial environments remains limited. The main challenges are their high computational cost, memory requirements, and energy consumption. While individual efficiency techniques such as neural architecture search (NAS), token compression, and low-precision inference have been extensively studied, most prior work targets only a single optimization axis, limiting overall deployment gains while preserving accuracy. 
In this paper, we present one of the first holistic frameworks that jointly optimizes three complementary axes: architecture, token, and bit-width. Specifically, the framework identifies compact backbones via Neural Architecture Search (AutoFormer), reduces information processing via token merging (ToMe), and accelerates per-operation execution via fp16 mixed-precision inference. 
In our study, we analyze accuracy-efficiency trade-offs on ImageNet-1K under aggressive compression. We then apply the selected configurations to a real-world in-house 3D X-ray semiconductor defect classification dataset for IC chip packaging inspection.
Results show that the proposed multi-axis framework achieves more than $10\times$ improvement in throughput along with over $10\times$ reductions in parameter count, FLOPs, and energy consumption, while maintaining the required accuracy on the downstream industrial task. To the best of our knowledge, this is among the earliest works to jointly optimize architecture, token, and bit-width dimensions in ViTs and the first such resource-efficient, deployment-focused study tailored to semiconductor manufacturing.
\end{abstract}


\section{Introduction}

Vision Transformers (ViTs)~\cite{vit} have achieved strong visual recognition performance, but their deployment remains challenging due to their high computational cost, memory demand, and energy consumption. In practical settings, these limitations become especially significant when inference must satisfy constraints on model size, runtime efficiency, and power usage.

\textbf{Research Gap.}
Recent studies have explored several directions for efficient ViTs~\cite{saha2025vision}, including compact backbone design via NAS~\cite{autoformer,gong2022nasvit}, token compression~\cite{DyViT_2021,tome}, and reduced-precision inference~\cite{yuan2022ptq4vit,li2023vit}.
However, most existing efforts address only a single optimization axis in isolation. Consequently, the complementary benefits of jointly optimizing architecture-level, token-level, and bit-width dimensions remain largely unexplored.
Moreover, the majority of prior work is evaluated primarily on standard benchmarks such as ImageNet-1K, with limited attention to real-world industrial applications and rigorous hardware-aware metrics such as throughput and energy consumption. In particular, there is a notable gap in holistic frameworks that systematically combine multiple optimization axes for practical deployment, especially in specialized and resource-constrained domains such as semiconductor manufacturing inspection.

\textbf{Our contribution.}
To address these limitations, we propose a holistic framework that jointly optimizes three complementary axes: architecture, token, and bit-width for Vision Transformer deployment. Specifically, the framework identifies compact backbones via Neural Architecture Search (AutoFormer), reduces token-level computation via token merging (ToMe), and accelerates per-operation execution via fp16 mixed-precision inference. Starting from a DeiT-B/16 baseline~\cite{deit}, we first analyze accuracy-efficiency trade-offs on ImageNet-1K under aggressive compression. The selected configurations are then transferred and fine-tuned on a real-world in-house 3D X-ray semiconductor defect classification dataset for IC chip packaging inspection. Experimental results demonstrate that the proposed multi-axis framework achieves more than $10\times$ improvement in throughput together with over $10\times$ reductions in parameter count, FLOPs, and energy consumption, while maintaining the required classification accuracy on the downstream industrial task. To the best of our knowledge, this is among the earliest works to jointly optimize architecture, token, and bit-width dimensions in Vision Transformers and the first such deployment-focused study tailored to semiconductor manufacturing.

The remainder of the paper is organized as follows. Section~\ref{sec:related_work} reviews prior work on compact Vision Transformers, token compression, and model quantization. Section~\ref{sec:method} presents the proposed framework. Section~\ref{sec:experiments} describes the experimental setup and evaluation protocol, and reports the main experiments and 3D X-ray semiconductor defect classification experiments. Section~\ref{sec:conclusion} concludes the work.

\section{Related Work}
\label{sec:related_work}

\subsection{Token Compression}
Vision Transformers process images as sequences of visual tokens, but not all tokens contribute equally to the final prediction. For example, background regions or highly redundant local patterns may contain limited semantic information~\cite{tcformer}. Token compression methods exploit this property to reduce the sequence-processing cost of ViTs by decreasing the number of tokens that remain active during inference.

Existing token compression methods are commonly grouped into two main families: token pruning and token merging. Token pruning estimates token importance and discards less useful tokens in selected layers, often through lightweight scoring modules or data-dependent keep policies~\cite{EViT_2022, DyViT_2021, SPViT_2022}. Token merging, in contrast, reduces sequence length by combining similar tokens, aiming to preserve information while lowering computational cost. Token compression is attractive because it can often be applied on top of an existing transformer without redesigning the full architecture.

In particular, we focus on token merging within NAS-based compact ViTs and analyze whether this additional compression stage can further improve the deployment trade-off between accuracy and efficiency.

\subsection{Structure Compression}
To enable efficient deployment of Vision Transformers (ViTs) in computation-intensive applications~\cite{abdollahzadeh2021revisit,mentzer2022vct,cheung2009highly} and resource-constrained settings such as edge AI~\cite{tran2018device,sander2025accelerating} and AI-agent systems~\cite{zhou2024survey,chen2025rlrc}, structure compression has become an important research direction. 
It aims to reduce redundant computation and model size while preserving predictive performance. 
Existing structure compression methods can be broadly grouped into two categories: pruning-based approaches, which remove redundant structural components from a given model, and Neural Architecture Search (NAS), which searches for compact architectures under accuracy-efficiency constraints.

Pruning-based methods introduce sparsity by eliminating unimportant weights, channels, or layers, thereby reducing computational cost and latency. For Vision Transformers, representative examples include channel pruning in NViT~\cite{yang2021nvit} and width/depth pruning in WDPruning~\cite{wdpruning}. Although effective, pruning-based pipelines often require iterative pruning and fine-tuning, which makes it costly to obtain and compare multiple compact architectures.

NAS provides a convenient alternative for generating architectures with different efficiency levels. AutoFormer~\cite{autoformer} proposes a one-shot NAS framework in which a weight-sharing supernet is trained once and compact subnetworks are later selected through evolutionary search. This paradigm is particularly attractive for deployment studies because it enables convenient access to multiple compact architectures with different accuracy-efficiency trade-offs. Follow-up works have further extended this direction by enlarging the search space~\cite{autoformer_v2} or improving supernet training and transferability, as in NasViT~\cite{gong2022nasvit} and ViTAS~\cite{su2022vitas}.

In this work, we adopt AutoFormer subnets as compact ViT backbones. This choice is motivated not only by their strong reported efficiency, but also by the practicality of NAS for obtaining multiple compact architectures without the extensive retraining cost often required by structural pruning.

\subsection{Quantization}
Quantization is one of the effective ways to compress Vision Transformers (ViTs). Instead of using 32-bit floating-point representation, parameters can be quantized to lower bit-widths (e.g., 8-bit or even 4-bit) to reduce the model size and the computational complexity. Besides, Post-Training Quantization (PTQ) does not require re-training or fine-tuning, which is very time-consuming. Thus, it is efficient and feasible even when the training dataset is not available. Recently, several PTQ methods have been proposed for ViTs, including PTQ~\cite{liu2021post}, PTQ4ViT~\cite{yuan2021ptq4vit}, APQ-ViT~\cite{ding2022towards}, FQ-ViT~\cite{lin2022fqvit}, NoisyQuant~\cite{liu2023noisyquant}, and I-ViT~\cite{li2023vit}.

\section{Proposed Method}
\label{sec:method}
We develop a multi-axis compression framework for practical Vision Transformer deployment. Starting from DeiT-B/16~\cite{deit} as the reference baseline, we jointly reduce architecture-level model size, token-level computation, and bit-width-level execution cost while preserving classification accuracy for the industrial semiconductor IC packaging application.
To evaluate the accuracy-efficiency trade-off, we consider multiple metrics, including classification accuracy, parameter count, GFLOPs, throughput, and energy consumption. Accuracy reflects downstream utility, while parameter count and GFLOPs characterize model compactness and nominal computational cost. Throughput and energy consumption capture practical deployment behavior on hardware.


To reduce model-level complexity, we replace the baseline backbone with compact architectures discovered by AutoFormer~\cite{autoformer}. AutoFormer is a one-shot Neural Architecture Search (NAS) framework that searches transformer design dimensions such as depth, embedding width, and attention structure, while avoiding the need to retrain every candidate architecture from scratch. It is particularly attractive in our setting because its weight-sharing supernet training allows sampled subnetworks to inherit well-optimized weights, improving the practical usability of the searched backbones and providing them strong initialization weights for downstream transfer learning. In this study, we adopt AutoFormer-T and AutoFormer-S as compact backbones to reduce parameter count and computational cost relative to DeiT-B/16.


After selecting compact backbones, we further reduce inference cost through token compression. 
In image classification, many visual tokens may correspond to background regions or redundant local patterns, and therefore contribute less to the final prediction. 
Reducing computation on such redundant tokens can improve efficiency while preserving accuracy.
Particularly, we 
apply Token Merging (ToMe)~\cite{tome} on top of the compact backbones. ToMe progressively merges similar tokens during inference, reducing the effective sequence length in later transformer layers and thereby lowering computation. 
A key advantage of ToMe is that it can be applied to an existing transformer architecture without redesigning the backbone itself. This makes it well suited to our setting, where the goal is not to develop a new architecture but to examine whether token compression remains beneficial after architecture-level compression. The combination of AutoFormer and ToMe is central to our method: AutoFormer reduces the model complexity at the backbone level, while ToMe further reduces the sequence-processing cost within that backbone. Since the two techniques act on different sources of inference cost, they are expected to be complementary.


Subsequently, the model is compressed under fp16 mixed-precision inference setting for practical deployment.
This design is motivated by the observation that architecture-level, token-level, and bit-width-level compression address different bottlenecks of Vision Transformer inference. The experiments in the following sections evaluate this design on both ImageNet-1K and the in-house 3D X-ray semiconductor defect classification.

\section{Experiments}
\label{sec:experiments}

\begin{table}[!t]
\centering
\caption{ImageNet-1K analysis used for pipeline selection. Here, $\mathrm{AutoF\mbox{-}S}$ and $\mathrm{AutoF\mbox{-}T}$ denote the AutoFormer-Small and AutoFormer-Tiny backbones, respectively, and the subscript $r$ denotes the ToMe token merging ratio.}
\label{tab:imagenet1k_results}
\setlength{\tabcolsep}{4pt}
\footnotesize
\begin{tabular}{|l|c|c|c|c|c|}
\hline
\textbf{Model} & \textbf{Acc.}$\uparrow$ & \textbf{FPS}$\uparrow$ & \textbf{Energy (J)}$\downarrow$ & \textbf{\#params}$\downarrow$ & \textbf{GFLOPs}$\downarrow$ \\
\hline
baseline                         & 81.8 & 34.1  & 5969.2 & 85.8M & 16.9 \\
$\mathrm{AutoF\mbox{-}S}$        & 81.7 & 153.6 & 1236.9 & 22.5M & 5.2 \\
$\mathrm{AutoF\mbox{-}S}_{r=10}$ & 78.3 & 179.6 & 935.8  & 22.5M & 3.3 \\
$\mathrm{AutoF\mbox{-}S}_{r=15}$ & 77.1 & 223.1 & 726.8  & 22.5M & 2.4 \\
$\mathrm{AutoF\mbox{-}T}$        & 75.3 & 282   & 548.3  & 5.7M  & 1.5 \\
$\mathrm{AutoF\mbox{-}T}_{r=10}$ & 73.9 & 227.7 & 594.0  & 5.7M  & 0.9 \\
$\mathrm{AutoF\mbox{-}T}_{r=15}$ & 72.6 & 330.2 & 424.2  & 5.7M  & 0.7 \\
\hline
\end{tabular}
\end{table}

\begin{table}[!t]
\centering
\caption{Deployment results on the 3D X-ray semiconductor defect dataset after fine-tuning the selected compressed variants. Here, $\mathrm{AutoF\mbox{-}S}$ and $\mathrm{AutoF\mbox{-}T}$ denote AutoFormer-Small and AutoFormer-Tiny backbones, respectively, and $r$ is the ToMe token merging ratio. $\mathrm{AutoF\mbox{-}T}_{r=15}$ achieves the best efficiency, with over $10\times$ higher throughput and over $10\times$ reductions in parameters, FLOPs, and energy, while maintaining the required classification accuracy.}
\label{tab:wp5_results}
\setlength{\tabcolsep}{5pt}
\footnotesize
\begin{tabular}{|l|c|c|c|c|}
\hline
\textbf{Model} & \textbf{FPS}$\uparrow$ & \textbf{Energy (J)}$\downarrow$ & \textbf{\#params}$\downarrow$ & \textbf{GFLOPs}$\downarrow$ \\
\hline
baseline                         & 13.1  & 15339.6 & 85.8M & 16.9 \\
$\mathrm{AutoF\mbox{-}S}_{r=10}$ & 104.1 & 2321.5  & 22.5M & 3.3 \\
$\mathrm{AutoF\mbox{-}S}_{r=15}$ & 107.6 & 2238.4  & 22.5M & 2.4 \\
$\mathrm{AutoF\mbox{-}T}_{r=10}$ & 136.4 & 1472.9  & 5.7M  & 0.9 \\
$\mathrm{AutoF\mbox{-}T}_{r=15}$ & 161.1 & 1151.1  & 5.7M  & 0.7 \\
\hline
\end{tabular}

\end{table}

\subsection{Experimental Setup}
The experiments use two datasets with complementary roles. ImageNet-1K serves as the benchmark for selecting compact backbones (AutoFormer versions) and token compression settings under aggressive compression, and the selected variants are then transferred to the in-house 3D X-ray semiconductor defect dataset for deployment evaluation.

We evaluate a DeiT-B/16 baseline together with compact AutoFormer backbones and their ToMe-compressed variants. We denote the compressed models by $\mathrm{AutoF\mbox{-}S}_{r}$ and $\mathrm{AutoF\mbox{-}T}_{r}$, where $r$ is the ToMe merging ratio; for instance, $\mathrm{AutoF\mbox{-}S}_{r=10}$ denotes AutoFormer-Small with ToMe applied at merging ratio $r=10$. All ToMe-compressed AutoFormer variants are retrained for 100 epochs at $224\times224$ resolution using AdamW, weight decay 0.05, layer-wise learning rate decay 0.75, and fp16 mixed-precision training. The batch size is 1024 on ImageNet-1K, with the standard DeiT augmentation pipeline~\cite{deit}, and 1 for 3D X-ray semiconductor defect experiments.

For deployment, the retrained AutoFormer models are exported to ONNX with opset 17 and dynamic batch support, and then converted to TensorRT. We generate both FP16 and FP32 TensorRT engines and benchmark inference on an NVIDIA RTX 6000 Ada Generation GPU using TensorRT10.

On ImageNet-1K, we report both classification accuracy and efficiency metrics, including throughput (FPS), energy consumption (J), parameter count, and GFLOPs. On 3D X-ray semiconductor defect classification, all selected compressed variants are fine-tuned to recover the highest possible downstream accuracy on the test set (100\% accuracy), so we report only the efficiency gains on this dataset.

\subsection{3D X-ray semiconductor defect classification}
\begin{figure*}[!t]
\centering
\begin{subfigure}[t]{0.3\textwidth}
    \centering
    \includegraphics[height=3.0cm]{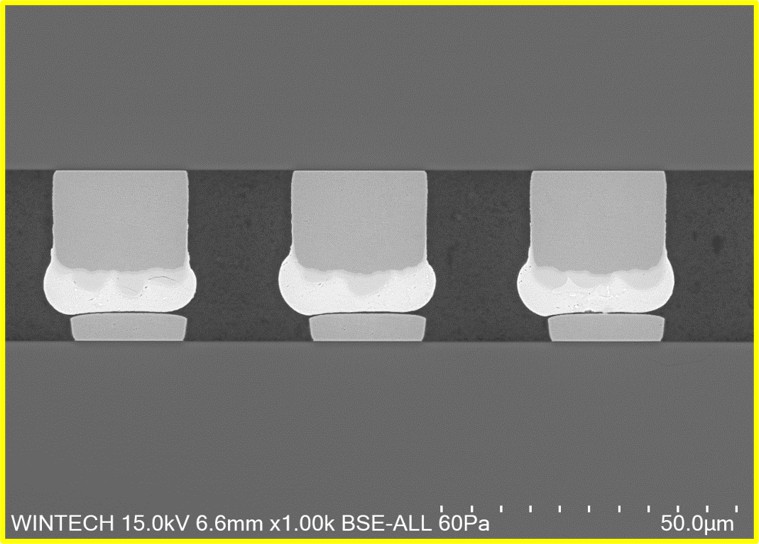}
    \caption{\textbf{Non-Wet.} Solder material did not wet the metal surface (UBM).}
    \label{fig:nonwet}
\end{subfigure}\hfill
\begin{subfigure}[t]{0.22\textwidth}
    \centering
    \includegraphics[height=3.0cm]{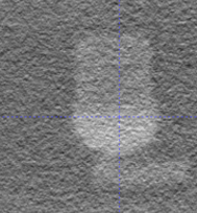}
    \caption{\textbf{Pad misalignment.} 
    The solder bump center 
    is offset from the metal pad center.}
    \label{fig:padmis}
\end{subfigure}\hfill
\begin{subfigure}[t]{0.30\textwidth}
    \centering
    \includegraphics[height=3.0cm]{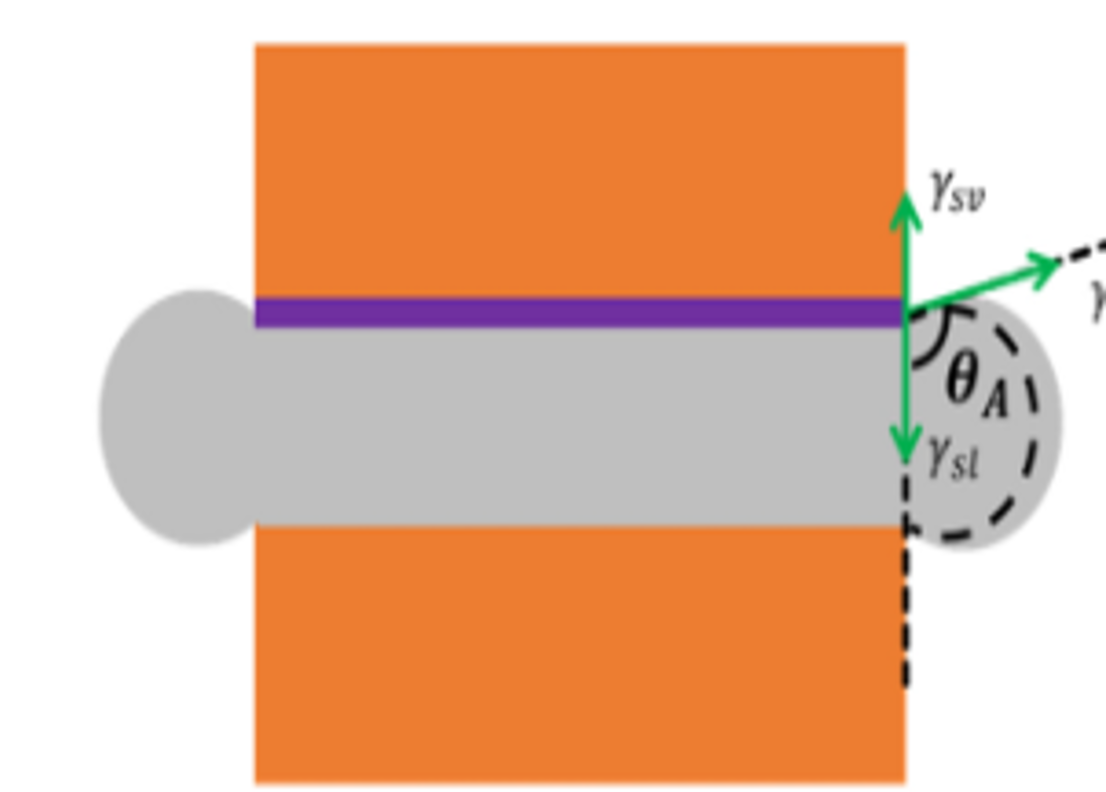}
    \caption{\textbf{Solder squeeze/extrusion.} Solder material is forced beyond the bump diameter.}
    \label{fig:squeeze}
\end{subfigure}


\begin{subfigure}[t]{0.6\textwidth}
    \centering
    \includegraphics[height=3.0cm]{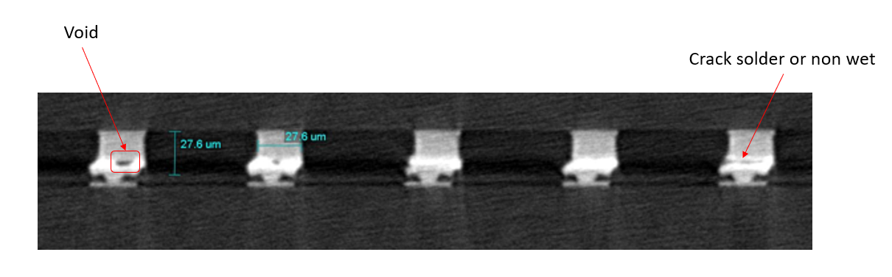}
    \caption{Void and Crack solder or non wet.}
    \label{fig:bottom}
\end{subfigure}
\caption{3D X-ray semiconductor defects. The subfigures show the appearances of different categories of defects.}
\label{fig:data}
\end{figure*}

We use a real-world, in-house 3D X-ray semiconductor defect dataset in our experiment, which captures various types of IC chip manufacturing defects that occur packaging buried inside the chip. The dataset used in this experiment consists of totally 93 3D volumes, of which 80 are used for training and 13 for testing. The maximum volume dimension for the data samples is around $[900, 900, 875]$ throughout the dataset. The dataset is fully labeled with ground truth defect classes ranging from 5 defect types, i.e., Pillar, Solder, Void and Pad, annotated manually by human experts. Fig.~\ref{fig:data} visualizes a dataset overview.

\subsection{ImageNet-1K Analysis and Pipeline Selection}
Table~\ref{tab:imagenet1k_results} reports the analysis  on ImageNet-1K. The purpose of this analysis is not to maximize benchmark accuracy, but to understand how compact backbones and token compression affect the accuracy-efficiency trade-off under aggressive compression.
Replacing the DeiT-B/16 baseline with AutoFormer already yields a strong improvement in efficiency. In particular, AutoFormer-S retains nearly the same top-1 accuracy as the baseline (81.7\% vs.\ 81.8\%) while reducing parameter count from 85.8M to 22.5M and GFLOPs from 16.9 to 5.2, with large gains in throughput and energy efficiency. AutoFormer-T pushes this trend further, reducing the model to 5.7M parameters and 1.5 GFLOPs, although with a larger drop in ImageNet-1K accuracy.

Applying token compression on top of the compact backbones further reduces computational cost. For both AutoFormer-S and AutoFormer-T, increasing the token merging ratio decreases GFLOPs and energy consumption and generally improves throughput. This confirms that token compression remains beneficial even after architecture-level compression. As expected, however, more aggressive token compression also leads to a larger accuracy drop on ImageNet-1K. The benchmark results therefore reveal a clear trade-off between efficiency and top-1 accuracy.

Based on these results, we select several compressed variants as deployment candidates for the in-house dataset. Rather than choosing models solely by benchmark accuracy, we prioritize variants that provide strong efficiency gains while remaining within a reasonable accuracy range under aggressive compression. These selected models are then fine-tuned on 3D X-ray semiconductor defect classification to assess whether the efficiency gains can be retained while recovering downstream task performance.

\subsection{Deployment Results on 3D X-ray Semiconductor Defect Classification}
Table~\ref{tab:wp5_results} reports the deployment results. Unlike the ImageNet-1K benchmark, the goal here is not to compare raw accuracy across unadapted models. Instead, all selected compressed variants are fine-tuned on the in-house dataset to recover the highest possible downstream classification performance. For this reason, the in-house comparison focuses on efficiency improvement after adaptation.

The results show that the efficiency gains observed on ImageNet-1K transfer effectively to the target application. All compressed variants substantially outperform the baseline in throughput, energy consumption, parameter count, and GFLOPs. In particular, the AutoFormer-S variants already provide a large deployment advantage over the baseline, while the AutoFormer-T variants deliver even stronger efficiency gains due to their more compact backbone.

Among the evaluated pipelines, $\mathrm{AutoF\mbox{-}T}_{r=15}$ provides the strongest deployment result. Compared with the baseline, it increases throughput from 13.1 to 161.1 FPS, corresponding to a speedup of about $12.3\times$, while reducing energy consumption from 15339.6~J to 1151.1~J, which is about $13\times$ lower energy. 
At the same time, it reduces model size from 85.8M to 5.7M parameters, corresponding to a $15.1\times$ smaller model, and lowers the computation budget from 16.9 to 0.7 GFLOPs, which is a $24.1\times$ reduction. These results show that the selected pipeline not only satisfies but exceeds the target of $10\times$ deployment improvement on the downstream task.

Overall, the in-house experiments confirm the main hypothesis of this work: combining compact backbone selection with token compression produces a practical deployment-ready ViT for image classification. Benchmark experiments on ImageNet-1K are useful for identifying favorable accuracy-efficiency operating points, while fine-tuning on the target dataset allows these compressed models to retain task performance and realize substantial deployment gains in practice.

\section{Conclusion}
\label{sec:conclusion}

This paper presented a practical study on improving Vision Transformer deployment efficiency for image classification. Using compact AutoFormer backbones together with token compression (ToMe), we identified compressed model variants that maintain favorable accuracy-efficiency trade-offs on ImageNet-1K and provide substantial efficiency gains on the in-house 3D X-ray semiconductor defect classification task. Overall, the study shows that multi-axis optimization of ViT combining compact backbone design with token compression and bit-width reduction can be an effective and practical approach for resource-efficient, deployment-oriented ViT optimization for 3D X-ray semiconductor defect classification. Future work extends the approach for other tasks
\cite{song2016egocentric,min2021integrating,chandrasegaran2022discovering}.

\section*{Acknowledgment}
This research is supported by Temasek Laboratories, Singapore University of Technology and Design; the National Research Foundation, Singapore under its AI Singapore Programmes (AISG Award No.: AISG2-TC-2022-007); the Agency for Science, Technology and Research (A*STAR) under its MTC Programmatic Funds (Grant No. M23L7b0021); the National Research Foundation, Singapore and Infocomm Media Development Authority under its Trust Tech Funding Initiative. Any opinions, findings and conclusions or recommendations expressed in this material are those of the author(s) and do not reflect the views of National Research Foundation, Singapore and Infocomm Media Development Authority.

\bibliographystyle{IEEEtran}
\bibliography{main}

\end{document}